\documentclass[10pt,twocolumn,letterpaper]{article}

\usepackage{cvpr}              

\definecolor{cvprblue}{rgb}{0.21,0.49,0.74}
\usepackage[pagebackref,breaklinks,colorlinks,allcolors=cvprblue]{hyperref}

\usepackage{graphicx}
\usepackage{multirow}
\usepackage{booktabs}

\newcommand*\rot{\rotatebox{90}}

\usepackage{colortbl}

\usepackage[T1]{fontenc}
\usepackage{makecell}
\usepackage{rotating}

\usepackage[absolute,overlay]{textpos}

\def\paperID{56} 
\def\confName{CVPR}
\def\confYear{2025}

\title{Improving Wildlife Out-of-Distribution Detection: Africa's Big Five}

\author{Mufhumudzi Muthivhi\\
Institute for Artificial Intelligent Systems\\
University of Johannesburg\\
{\tt\small mmuthivhi@uj.ac.za}
\and
Jiahao Huo\\
Institute for Artificial Intelligent Systems\\
University of Johannesburg\\
{\tt\small jhuo@uj.ac.za}
\and
Fredrik Gustafsson\\
Department of Electrical Engineering\\
Linköping University\\
{\tt\small fredrik.gustafsson@liu.se}
\and
Terence L. van Zyl\\
Institute for Artificial Intelligent Systems\\
University of Johannesburg\\
{\tt\small tvanzyl@uj.ac.za}
}

\begin{document}
\maketitle
\begin{abstract}
Mitigating human-wildlife conflict seeks to resolve unwanted encounters between these parties. Computer Vision provides a solution to identifying individuals that might escalate into conflict, such as members of the Big Five African animals.
However, environments often contain several varied species. The current state-of-the-art animal classification models are trained under a closed-world assumption. They almost always remain overconfident in their predictions even when presented with unknown classes.
This study investigates out-of-distribution (OOD) detection of wildlife, specifically the ``Big Five''.
To this end, we select a parametric Nearest Class Mean (NCM) and a non-parametric contrastive learning approach as baselines to take advantage of pretrained and projected features from popular classification encoders. Moreover, we compare our baselines to various common OOD methods in the literature.
The results show feature-based methods reflect stronger generalisation capability across varying classification thresholds. Specifically, NCM with ImageNet pretrained features achieves a $2\%$, $4\%$ and $22\%$ improvement on AUPR-IN, AUPR-OUT and AUTC over the best OOD methods, respectively. The code can be found here \href{https://github.com/pxpana/BIG5OOD}{https://github.com/pxpana/BIG5OOD}
\end{abstract}    
\begin{textblock*}{\textwidth}(2cm,1cm)
  \small \centering \textit{Accepted to the CVPR 2025 Workshop on Computer Vision for Animal Behavior Tracking and Modeling (CV4Animals)}
\end{textblock*}

\section{Introduction}
\label{sec:intro}

Human-wildlife conflict results from unwanted interactions between humans, livestock and wildlife. Communities living alongside wildlife sanctuaries often find themselves competing for resources with the wildlife~\cite{braczkowski2023unequal}. For instance, livestock predation and crop raiding can escalate to retaliatory killings, hurting conservation efforts. The buffalo, elephant, lion, leopard and rhino are considered Africa's flagship species~\cite{caro2010conservation,williams2000flagship}. They are often referred to as the Big Five and are a major tourist attraction. They also exhibit aggressive behaviour, potential for property damage, frequent encroachment into human settlements and entrapment~\cite{di2021pan,dunham2010human,sedhain2016human}. Their hostile nature further threatens their endangered status. To mitigate human-wildlife interactions, common strategies include electric fencing, artificial repellents, and motion sensors~\cite{janani2022human}. Recent literature explores the use of edge devices to classify wild animals and simultaneously repel them using a buzzer or electronic firecrackers~\cite{sambhaji2019early,bavane2018protection,mishra2024smart}. 

Meanwhile, large-scale classification models have succeeded in wildlife monitoring and ecological studies~\cite{berger2017wildbook}. The largest of which, SpeciesNet, can classify up to $2000$ animals. These state-of-the-art ``foundational'' models are trained under the closed-world assumption~\cite{villa2017towards, swanson2015snapshot, willi2019identifying, stevens2024bioclip, Beery_Efficient_Pipeline_for, gadot2024crop}. They perform well in the classes they have seen during training, but will misclassify unknown classes as known classes~\cite{yang2024generalized}. Therefore, practically, a model would have to be trained on every species in a region, even if the model is intended for use on a limited subset. 

This paper explores the training of a Big Five classification model. We handle false detections by labelling any animals not part of the big five as out-of-distribution (OOD). There is limited recent research on using machine learning to monitor the Big Five African iconic animals using camera trap data. Further, OOD detection remains an open research challenge, and even fewer studies are exploring how these methods perform on OOD wildlife data beyond the benchmarks.
This study evaluates how well current state-of-the-art (SOTA) pre-trained architectures perform on the Big Five animal classification. Then we conduct several experiments on existing inference and feature-based OOD methods to detect animals not included in the training data~\cite{kirchheim2022pytorch}.
The results show feature-based methods maintain the strongest performance for AUROC, AUPR-IN, AUPR-OUT and AUTC metrics. Our selected baseline methods approach current SOTA in OOD detection and motivate continued research in feature-based methods for OOD wildlife monitoring.
We contribute to the literature by:
\begin{enumerate}
    \item providing a Big Five classification and OOD detection model using ImageNet features;
    \item establishing a simple but effective agreement-based dual-head prediction algorithm;
    \item demonstrating the superiority of general-purpose pre-trained features for detecting OOD samples; and
    \item conducting a comparison against state-of-the-art OOD methods for each pretrained classification model.
\end{enumerate}

\section{Background}
\label{sec:background}


There is increasing interest, across various ecosystems, in using computer vision for recognising different wildlife species from images and videos~\cite {saoud2024beyond}. \citet{villa2017towards} uses camera-trap images with a Neural Network to classify animals from the Snapshot Serengeti dataset. \citet{willi2019identifying} propose using citizen science to generate annotations from camera traps in the Serengeti. MegaClassifier uses cropped MegaDetector images from predominantly North American and European species~\cite{Beery_Efficient_Pipeline_for}. BioClip is trained on the TreeOfLife-10M dataset, which encompasses various animals, plants, fungi and insects~\cite{stevens2024bioclip}. SpeciesNet is trained as an ensemble with MegaDetector on over $60$ million images with approximately $2,000$ classes. However, none of these methods can detect animals outside their distribution. 

\subsection{Out-of-distribution (OOD) Detection}
Several strategies have been proposed to distinguish between known and unknown classes~\cite{yang2024generalized}. These approaches can be divided into inference and feature regularisation-based methods~\cite{roady2020open}. Inference methods use pre-trained models while modifying the prediction outputs. The earliest of these methods threshold the maximum logits or softmax probabilities and are regarded as a general baseline in OOD~\cite{hendrycks2017a, pmlr-v162-hendrycks22a}. Temperature scaling has proven effective at calibrating predictions~\cite{guo2017calibration, liang2017enhancing}. Energy-based methods capture the overall uncertainty in the model’s predictions~\cite{liu2020energy, du2022vos}. Other strategies adopt a distance metric with a nearest neighbour clustering algorithm over the feature space~\cite{lee2018simple,sun2022out}. Extreme Value Theory methods record the occurrences of probabilities that are outliers~\cite{bendale2016towards, rudd2017extreme}. Feature regularisation methods modify the network to learn OOD representations. They can further be divided into unsupervised and supervised approaches. Unsupervised methods only use the in-distribution (ID) data during training. Deep Support Vector Data Description (SVDD) places a single class prototype in the embedding spacer~\cite{ruff2018deep}. Center loss extends the idea by assigning a prototype to each class~\cite{wen2016discriminative}. The objective is to pull all the representations of ID data closer to its associated class prototype. Recent methods are aimed at producing a confidence score given the feature space~\cite{lee2018simple, sehwag2021ssd, sun2022out}. Subsequent work fit class-conditional Gaussian distributions over the feature space activations to derive a discriminative decision boundary between ID and OOD samples~\cite{vojivr2023calibrated, winkens2020contrastive}. In particular, \citet{vojivr2023calibrated} leverages the Nearest Class Mean (NCM) to create class-specific feature prototypes. \citet{winkens2020contrastive} use contrastive learning to learn a label-agnostic feature space that is more discriminative. The supervised methods require OOD data during training. In the context of wildlife learning, such data is hard to obtain. Hence, this study only trains on ID data and evaluates both ID and OOD. 
\section{Methodology}
\label{sec:methodology}

\subsection{Methods}

We explore four pre-trained backbone encoders $f$: SpeciesNet, MegaClassifier, BioClip, and a Vision Transformer base (ViT) pre-trained on Imagenet~\cite{gadot2024crop, Beery_Efficient_Pipeline_for, stevens2024bioclip, caron2021emerging}. Given an image $x$, we extract its features $z = f(x)$ by freezing $f$. We train a two-layer classification head $g$ separated with a ReLU to obtain a prediction $y_1 \in \mathbb{R}^{5}$. We explore a parametric and non-parametric approach to determine if $x$ is in-distribution (ID) or out-of-distribution (OOD).

\subsubsection{Nearest Class Mean}
For our proposed parametric approach, we use Nearest Class Mean (NCM) to compute the average feature vector $\mu \in \mathbb{R}^{d_1}$ for the class $c$ in the validation set $\mathcal{D}$, such that:
\begin{equation}
    \mu_c = \frac{1}{| \mathcal{D}_c |} \sum_{x \in \mathcal{D}_c} f(x)
\end{equation}
where $d$ is the feature dimension. We classify the image $x$ by finding the NCM in the feature space such that:
$    y_2 = \arg\!\min_{c} \| f(x) - \mu_c \|_2 $.
An image $x$ is ID if $y_1 = y_2$. When the classification head and the NCM agree on the predicted class, we assume it is ID; otherwise, it is OOD.

\subsubsection{Contrastive Learning with KNN}
Our non-parametric approach concurrently trains a classification $g$ and a projection head $p$. Given the projected features $\hat{z} = p(z)$, we use a contrastive learning loss function, Normalized Temperature-scaled Cross Entropy (NTXent), to learn a discriminative feature space such that:
\begin{equation}
    - \log \frac{\exp(\mathrm{sim}(\hat{z}_i, \hat{z}_j)/\tau}{\sum_{k \neq 1}^{2N} \mathbb{1}_{k \neq i} \exp(\mathrm{sim}(\hat{z}^{A}_{i}, \hat{z}_{k})/\tau)}
\end{equation}
where $\hat{z}_{i}$ and $\hat{z}_j$ are the positive pairs for a batch of size $N$ and temperature term $\tau$~\cite{chen2020simple}. For OOD detection, we use $k$-nearest neighbours in the projected space. We run KNN (k=50) over the validation set, and select the majority class $y_2$ for an image $x$. Like NCM, $x$ is ID if the classification head and KNN predict the same class.

\subsection{Experimental Setup}
We adopt common OOD benchmarks from the PyTorch-OOD library~\cite{kirchheim2022pytorch}. We use Pytorch Lightning to streamline our training and testing process~\cite{pytorch2019lightning}. Each backbone architecture is trained with a classification and/or projection head. We use the Weighted Adam Optimiser with a linearly scaled learning rate of $0.005$ and a cosine warmup schedule. We limit all training to $100$ epochs, monitoring the validation performance.

\subsubsection{Dataset}
\begin{table}[htb!]
    \centering
    \caption{Number of images for In- and Out-of-Distribution Classes}
    \resizebox{0.8\columnwidth}{!}{%
    \begin{tabular}{lr|lr}
        \toprule
        \multicolumn{2}{c|}{In-Distribution} & \multicolumn{2}{c}{Out-of-Distribution} \\
        Species   & Images \# & Species     & Images \# \\
        \bottomrule \toprule
        Buffalo  & $48,184$      & Impala      & \multirow{6}{*}{\rot{$5,000$ for each}} \\
        Elephant & $54,453$      & Zebra       & \\
        Lion     & $15,584$      & Cheetah     & \\
        Leopard  & $6,487$       & Giraffe     & \\
        Rhino    & $7,434$       & Wildebeest  & \\
                 &               & Hippo      & \\
        \midrule
        Training   &  $79,285$    &    -        & - \\
        Validation &  $26,428$    & Validation  & $9,000$ \\
        Test       &  $26,428$    & Test        & $21,000$ \\
        \bottomrule
    \end{tabular}
    }    
    \label{tab:animal_images}
\end{table} 

We use several wildlife datasets available on LILA BC biology to create a subset of the African Big Five~\cite{lila2024science}. We employed Megadetector to crop the animals from the images, using a confidence of $0.2$. We use K-means clustering to group similar images together to avoid overlapping clusters between the train-test split. We employ the same method to acquire six additional species for OOD testing. Table \ref{tab:animal_images} shows the ID and OOD animals used in our experiments. These OOD species are selected for their visual similarity to the ID classes as a challenging open-world problem. For example, the patterns on the giraffe and cheetah are similar to those of leopards under certain lighting conditions in camera traps. Similarly, wildebeest, impala and hippos have body shapes that could be similar to buffalo, lion and rhino under varying conditions, respectively. Elephants and zebras frequently co-occur in the same habitats, leading to potential information leakage in training data. Table \ref{tab:animal_images} shows the counts of each species. The training, validation, and testing splits were $60\%$, $20\%$, and $20\%$, respectively. The splits were stratified by class labels to preserve the ratios between classes. When evaluated on accuracy metrics, we take $30\%$ of images from the OOD classes as a validation set to find the optimal thresholds for the non-parametric methods. The images are resized and cropped to $224\times 224$. 

\subsubsection{Metrics}
\label{sec:metrics}
We adopt four commonly used OOD metrics. Area Under the Receiver Operating Characteristic Curve (AUROC) measures the models' ability to distinguish between ID and OOD samples over all possible classification thresholds~\cite{fawcett2006introduction}. An AUROC of one represents a perfect model, whereas $0.5$ is a random classifier. Area Under the Precision-Recall Curve (AUPR) examines the trade-off between precision and recall~\cite{powers2020evaluation}. AUPR-IN treats ID samples as the positive class, whereas AUPR-OUT favours OOD samples. Area Under the Threshold Curve (AUTC) avoids binarising OOD scores and instead considers the distribution of ID and OOD samples by penalizing poor separation between them~\cite{humblot2023beyond}.

\subsubsection{Pretrained Backbones and  OOD Methods}
We tested a trainable linear layer baseline after the various backbone architectures and, secondly, one non-linear and one linear layer. Thirdly, we employed strong augmentations on the rhinos' images due to the limited availability of images. We selected a non-linear baseline with strong augmentations and oversampling of rhinos due to its stronger performance on per-class accuracy metrics.
We employed twelve existing OOD solutions consisting of parametric and non-parametric methods. The feature-based methods include DeepSVDD and Center Loss, and the inference-based methods are MaxSoftmax, EnergyBased and MaxLogit. We compare these against our proposed two methods as new baselines.

\begin{table}[htb!]
\caption{Classification performance of finetuned pretrained backbones on the in-distribution species}
\centering
\resizebox{0.8\columnwidth}{!}{
\begin{tabular}{lrrrr}
\toprule
\textit{\textbf{Species}} 
& \rot{\makecell[l]{Species-\\Net}} & \rot{\makecell[l]{Mega-\\Classifier}} & \rot{BioClip} & \rot{ImageNet} \\
\midrule
\toprule
{Elephant}  & $.475$ & $.884$ & $.827$ & $.904$ \\ 
{Buffalo}   & $.799$ & $.882$ & $.889$ & $.951$ \\ 
{Lion}      & $.084$ & $.843$ & $.761$ & $.905$ \\ 
{Leopard}   & $.322$ & $.902$ & $.939$ & $.958$ \\ 
{Rhino}     & $.114$ & $.294$ & $.335$ & $.529$ \\
\midrule
\textbf{F1 Score \tiny{(Macro Ave.)}}     & $.376$ & $.749$ & $.761$ & $.862$ \\
\textbf{F1 Score \tiny{(Weighted Ave.)}}  & $.487$ & $.827$ & $.815$ & $.900$ \\
\bottomrule
\end{tabular}}
\label{table:accuracy}
\end{table}

\def\ccb{\cellcolor{blue!20}}
\def\ccB{\cellcolor{blue!20}\textbf}

\begin{table*}[htb!]
\caption{Out-of-Distribution performance of each method. Top three methods per model are highlighted. Best overall in bold.}
\centering
\resizebox{\textwidth}{!}{%
\begin{tabular}{lrrrr|rrrr|rrrr|rrrr}
\toprule
\textit{\textbf{Metrics}} & 
    \multicolumn{4}{c}{\textbf{AUROC} $\uparrow$} & 
    \multicolumn{4}{c}{\textbf{AUPR-IN} $\uparrow$} & 
    \multicolumn{4}{c}{\textbf{AUPR-OUT} $\uparrow$} &
    \multicolumn{4}{c}{\textbf{AUTC} $\downarrow$ } \\
\cmidrule(lr){2-5} \cmidrule(lr){6-9} \cmidrule(lr){10-13} \cmidrule(lr){14-17} \\
\textit{\textbf{Models}} 
& \rot{\makecell[l]{Species-\\Net}} & \rot{\makecell[l]{Mega-\\Classifier}} & \rot{BioClip} & \rot{ImageNet}
& \rot{\makecell[l]{Species-\\Net}} & \rot{\makecell[l]{Mega-\\Classifier}} & \rot{BioClip} & \rot{ImageNet}
& \rot{\makecell[l]{Species-\\Net}} & \rot{\makecell[l]{Mega-\\Classifier}} & \rot{BioClip} & \rot{ImageNet}
& \rot{\makecell[l]{Species-\\Net}} & \rot{\makecell[l]{Mega-\\Classifier}} & \rot{BioClip} & \rot{ImageNet}\\
\bottomrule
\toprule
\textit{inference-based} \\
\; MaxSoftmax~\cite{hendrycks2017a}  & .523 & \ccb{.556} & .635 & .744 & .590 & .609 & .658 & .747 & .458 & .481 & .539 & .669 & .492 & .474 & .444 & .408  \\
\; MaxLogit~\cite{pmlr-v162-hendrycks22a}    & .536 & .541 & .683 & .750 & .589 & .591 & .675 & .701 & .478 & .472 & .635 & .700 & .499 & .496 & .478 & .463  \\
\; Temp'Scaling~\cite{guo2017calibration} & .475 & .478 & .445 & .540 & .564 & .550 & .520 & .582 & .444 & .420 & .417 & .464 & .500 & .500 & .454 & .496 \\
\; OpenMax~\cite{bendale2016towards}     & .526 & .530 & .663 & .724 & .574 & .579 & .673 & .719 & .473 & .566 & \ccb{.679} & .663 & .497 & .498 & .459 & .460 \\
\; KNN~\cite{sun2022out}         & \ccb{.557} & \ccB{.607} & .659 & .737 & .619 & .656 & .721 & .797 & .482 & .555 & .584 & .632 & .498 & .497 & .517 & .441 \\
\; EnergyBased~\cite{liu2020energy} & .537 & .531 & \ccb{.694} & .751 & .587 & .585 & .679 & .725 & .480 & .463 & .625 & \ccb{.705} & .499 & .497 & .480 & .465 \\
\; Entropy~\cite{shannon1948mathematical}     & .525 & .555 & .635 & .747 & .585 & .609 & .659 & .748 & .463 & .478 & .536 & .674 & .493 & .476 & .438 & .395 \\
\; SHE~\cite{zhang2022out}         & .528 & .500 & .526 & .494 & \ccb{.778} & \ccB{.779} & \ccB{.771} & .777 & .469 & \ccB{.721} & .454 & .412 & .498 & .502 & .498 & .424 \\
\; ReAct~\cite{sun2021react}       & .533 & .534 & \ccB{.695} & .732 & .584 & .584 & .681 & .733 & .477  & .466 & .658 & .667 & .499 & .497 & .476 & .469 \\
\; DICE~\cite{sun2022dice}        & .518 & .528 & .681 & .749 & .573 & .580 & .670 & .727 & .463 & .467 & .643 & .696 & .497 & .498 & .483 & .465 \\
\textit{feature-based} \\
\; Deep SVDD~\cite{ruff2018deep}   & .500 & .503 & .596 & .597 & \ccB{.779} & .563 & .642 & .634 & .571 & .441 & .548 & .559 & .500  & .500 & .499 & .499 \\
\; Center Loss~\cite{wen2016discriminative} & .531 & .514 & .647 & \ccb{.762} & .627 & .582 & .714 & .807 & \ccB{.712} & \ccb{.688} & \ccb{.709} & \ccB{.745} & .497 & .491 & .451 & .446 \\
\textit{inference and feature based} \\
\; GROOD~\cite{vojivr2023calibrated}         & .544 & \ccb{.597} & \ccb{.693} & \ccB{.824} & .603 & .645 & .573 & \ccB{.889} & .469 & .515 & \ccB{.764} & .682 & \ccb{.482} & \ccB{.461} & \ccb{.411} & \ccb{.322} \\
\midrule
\textit{proposed baselines} \\
\; \textbf{NCM Agreement}         & .518 & .531 & .594 & .693 & .654 & \ccb{.673} & \ccb{.747} & \ccb{.809} & \ccb{.632} & \ccb{.629} & .642 & \ccB{.734} & \ccB{.482} & \ccb{.469} & \ccB{.406} & \ccB{.306} \\
\; \textbf{Contrastive Loss} & \ccb{.547} & .546 & .580 & .634 & \ccb{.725} & \ccb{.698} & \ccb{.751} & .792 & \ccb{.591} & .621 & .619 & .683 & \ccb{.453} & \ccb{.454} & \ccb{.420} & \ccb{.367} \\
\; \textbf{NCM Agreement Score}*         & .542 & .532 & .651 & \ccb{.801} & .597 & .587 & .676 & \ccb{.858} & .477 & .467 & .554 & .700 & .490 & .492 & .475 & .408 \\
\; \textbf{Contrastive Agreement Score}*         & \ccB{.588} & .553 & .620 & .702 & .630 & .604 & .624 & .712 & .518 & .478 & .571 & .653 & .500 & .480 & .500 & .500 \\
\bottomrule
\multicolumn{7}{l}{* Recent results using the agreement score described in the supplementary material}
\end{tabular}
}
\label{table:OOD}
\end{table*}

\section{Results and Discussion}
\subsection{In-Distribution (ID)}
\label{sec:indistr}
The data in Table \ref{table:accuracy} shows that the ImageNet pre-trained backbone surpasses other models pre-trained on animal datasets when evaluating accuracy for all considered species. This result indicates that more generalised models across a wider range of animate and inanimate objects are more effective for OOD detection. Despite oversampling the Rhino data, each model achieved an accuracy below $60\%$ but on par with previous studies. The results of ID accuracy align with recent research on wildlife camera-trap-based classification~\cite{villa2017towards, 10.1145/3674029.3674050}. Our models significantly outperform these studies for buffalo, lions and leopards. For elephants, our results are on par with previous studies, having results between $85\%$ and $93.5\%$. 

\subsection{Out-of-Distribution (OOD)}
In Table \ref{table:OOD}, the results show that the best-performing method is our proposed NCM baseline using a pre-trained ImageNet backbone. Using the different OOD methods, the ImageNet backbone performs the best binary classification between OOD and ID classes. Interestingly, the results of backbones pretrained on wildlife species do not perform well on OOD detection compared to a more generalised model using ImageNet. 
EnergyBased with ImageNet produced the highest AUROC and Accuracy of $0.751$ and $0.728$, respectively, followed closely by MaxLogit. However, the highest results are achieved by NCM, Center Loss, SHE and ReAct when evaluating AUPR-IN and AUPR-OUT. This performance suggests that logit-based methods excel at global separability due to their direct modelling of class boundaries' uncertainty. By leveraging the embedding space, methods that use features achieve better precision-recall tradeoffs (AUPR) and can withstand imbalanced datasets.
The AUTC score for NCM and contrastive loss is the lowest at $0.306$ and $0.367$ with ImageNet, respectively. Moreover, AUTC does not view OOD as a binary classification problem. Instead, AUTC measures how effectively a model maintains separation between ID and OOD distributions across all possible decision thresholds. NCM's effectiveness stems from its class-conditional mean alignment. Contrastive learning is notable for its focus on optimising instance discrimination, as this creates robust decision boundaries that generalise well across various threshold choices. This performance advantage is especially significant when tackling real-world, multi-class classification problems, where optimal thresholds may be unknown or must adapt to changing data distributions. Table~\ref{table:BigFive} in the supplementary material reflects on the performance of the best-performing OOD models.

\section{Conclusion}\label{section:Conclusion}
This research aimed to evaluate and improve current state-of-the-art methods for detecting the African Big Five while minimising the misclassification of OOD individuals. Our findings indicate that pre-trained models using the ImageNet dataset outperform those trained on wildlife data for wildlife classification. Additionally, our OOD testing demonstrated that existing OOD detection methods perform better with an ImageNet-pretrained backbone. These results suggest that generalised features are more effective for wildlife detection. Furthermore, we proposed a straightforward baseline that utilises a multi-head output, incorporating the agreement of a classifier and a contrastive loss head.

{
    \small
    \bibliographystyle{ieeenat_fullname}
    \bibliography{main}

@String(CVPR= {IEEE Conf. Comput. Vis. Pattern Recog.})

@String(ICCV= {Int. Conf. Comput. Vis.})

@String(CVPR  = {CVPR})

@String(ICCV  = {ICCV})

@inproceedings{stevens2024bioclip,
      title = {{B}io{CLIP}: A Vision Foundation Model for the Tree of Life}, 
      author = {Samuel Stevens and Jiaman Wu and Matthew J Thompson and Elizabeth G Campolongo and Chan Hee Song and David Edward Carlyn and Li Dong and Wasila M Dahdul and Charles Stewart and Tanya Berger-Wolf and Wei-Lun Chao and Yu Su},
      booktitle={Proceedings of the IEEE/CVF Conference on Computer Vision and Pattern Recognition (CVPR)},
      year = {2024},
      pages = {19412-19424}
    }

@article{villa2017towards,
title       =   {Towards automatic wild animal monitoring: Identification of animal species in camera-trap images using very deep convolutional neural networks},
author      =   {Villa, Alexander Gomez and Salazar, Augusto and Vargas, Francisco},
journal     =   {Ecological informatics},
volume      =   {41},
pages       =   {24--32},
year        =   {2017},
publisher   =   {Elsevier}
}

@article{swanson2015snapshot,
title       =   {Snapshot Serengeti, high-frequency annotated camera trap images of 40 mammalian species in an African savanna},
author      =   {Swanson, Alexandra and Kosmala, Margaret and Lintott, Chris and Simpson, Robert and Smith, Arfon and Packer, Craig},
journal     =   {Scientific data},
volume      =   {2},
number      =   {1},
pages       =   {1--14},
year        =   {2015},
publisher   =   {Nature Publishing Group}
}

@article{willi2019identifying,
title       =   {Identifying animal species in camera trap images using deep learning and citizen science},
author      =   {Willi, Marco and Pitman, Ross T and Cardoso, Anabelle W and Locke, Christina and Swanson, Alexandra and Boyer, Amy and Veldthuis, Marten and Fortson, Lucy},
journal     =   {Methods in Ecology and Evolution},
volume      =   {10},
number      =   {1},
pages       =   {80--91},
year        =   {2019},
publisher   =   {Wiley Online Library}
}

@software{Beery_Efficient_Pipeline_for,
author      =   {Beery, Sara and Morris, Dan and Yang, Siyu},
license     =   {MIT},
title       =   {{Efficient Pipeline for Camera Trap Image Review}},
URL         =   {http://github.com/agentmorris/MegaDetector}
}

@article{gadot2024crop,
title       =   {To crop or not to crop: Comparing whole-image and cropped classification on a large dataset of camera trap images},
author      =   {Gadot, Tomer and Istrate, Ștefan and Kim, Hyungwon and Morris, Dan and Beery, Sara and Birch, Tanya and Ahumada, Jorge},
journal     =   {IET Computer Vision},
volume      =   {18},
number      =   {8},
pages       =   {1193--1208},
year        =   {2024},
publisher   =   {Wiley Online Library}
}

@article{yang2024generalized,
title       =   {Generalized out-of-distribution detection: A survey},
author      =   {Yang, Jingkang and Zhou, Kaiyang and Li, Yixuan and Liu, Ziwei},
journal     =   {International Journal of Computer Vision},
volume      =   {132},
number      =   {12},
pages       =   {5635--5662},
year        =   {2024},
publisher   =   {Springer}
}

@article{berger2017wildbook,
author = {Berger-Wolf, Tanya and Rubenstein, Daniel and Stewart, Charles and Holmberg, Jason and Parham, Jason and Menon, Sreejith and Crall, J.P. and Van Oast, Jon and Kiciman, Emre and Joppa, Lucas},
year = {2017},
month = {10},
pages = {},
title = {Wildbook: Crowdsourcing, computer vision, and data science for conservation},
doi = {10.48550/arXiv.1710.08880}
}

@article{saoud2024beyond,
title       =   {Beyond observation: Deep learning for animal behavior and ecological conservation},
author      =   {Saoud, Lyes Saad and Sultan, Atif and Elmezain, Mahmoud and Heshmat, Mohamed and Seneviratne, Lakmal and Hussain, Irfan},
journal     =   {Ecological Informatics},
pages       =   {102893},
year        =   {2024},
publisher   =   {Elsevier}
}

@article{roady2020open,
title       =   {Are open set classification methods effective on large-scale datasets?},
author      =   {Roady, Ryne and Hayes, Tyler L and Kemker, Ronald and Gonzales, Ayesha and Kanan, Christopher},
journal     =   {Plos one},
volume      =   {15},
number      =   {9},
pages       =   {e0238302},
year        =   {2020},
publisher   =   {Public Library of Science San Francisco, CA USA}
}

@inproceedings{
hendrycks2017a,
title={A Baseline for Detecting Misclassified and Out-of-Distribution Examples in Neural Networks},
author={Dan Hendrycks and Kevin Gimpel},
booktitle={International Conference on Learning Representations},
year={2017},
url={https://openreview.net/forum?id=Hkg4TI9xl}
}

@inproceedings{guo2017calibration,
title       =   {On calibration of modern neural networks},
author      =   {Guo, Chuan and Pleiss, Geoff and Sun, Yu and Weinberger, Kilian Q},
booktitle   =   {International conference on machine learning},
pages       =   {1321--1330},
year        =   {2017},
organization=   {PMLR}
}

@InProceedings{pmlr-v162-hendrycks22a,
  title = 	 {Scaling Out-of-Distribution Detection for Real-World Settings},
  author =       {Hendrycks, Dan and Basart, Steven and Mazeika, Mantas and Zou, Andy and Kwon, Joseph and Mostajabi, Mohammadreza and Steinhardt, Jacob and Song, Dawn},
  booktitle = 	 {Proceedings of the 39th International Conference on Machine Learning},
  pages = 	 {8759--8773},
  year = 	 {2022},
  editor = 	 {Chaudhuri, Kamalika and Jegelka, Stefanie and Song, Le and Szepesvari, Csaba and Niu, Gang and Sabato, Sivan},
  volume = 	 {162},
  series = 	 {Proceedings of Machine Learning Research},
  month = 	 {17--23 Jul},
  publisher =    {PMLR},
  url = 	 {https://proceedings.mlr.press/v162/hendrycks22a.html}
}

@article{lee2018simple,
title       =   {A simple unified framework for detecting out-of-distribution samples and adversarial attacks},
author      =   {Lee, Kimin and Lee, Kibok and Lee, Honglak and Shin, Jinwoo},
journal     =   {Advances in neural information processing systems},
volume      =   {31},
year        =   {2018}
}

@inproceedings{sun2022out,
title       =   {Out-of-distribution detection with deep nearest neighbors},
author      =   {Sun, Yiyou and Ming, Yifei and Zhu, Xiaojin and Li, Yixuan},
booktitle   =   {International Conference on Machine Learning},
pages       =   {20827--20840},
year        =   {2022},    
organization=   {PMLR}
}

@inproceedings{bendale2016towards,
title       =   {Towards open set deep networks},
author      =   {Bendale, Abhijit and Boult, Terrance E},
booktitle   =   {Proceedings of the IEEE conference on computer vision and pattern recognition},
pages       =   {1563--1572},
year        =   {2016}
}

@article{rudd2017extreme,
title       =   {The extreme value machine},
author      =   {Rudd, Ethan M and Jain, Lalit P and Scheirer, Walter J and Boult, Terrance E},
journal     =   {IEEE transactions on pattern analysis and machine intelligence},
volume      =   {40},
number      =   {3},
pages       =   {762--768},
year        =   {2017} ,
publisher   =   {IEEE}
}

@inproceedings{
liang2017enhancing,
title={Enhancing The Reliability of Out-of-distribution Image Detection in Neural Networks},
author={Shiyu Liang and Yixuan Li and R. Srikant},
booktitle={International Conference on Learning Representations},
year={2018},
url={https://openreview.net/forum?id=H1VGkIxRZ},
}

@article{liu2020energy,
title       =   {Energy-based out-of-distribution detection},
author      =   {Liu, Weitang and Wang, Xiaoyun and Owens, John and Li, Yixuan},
journal     =   {Advances in neural information processing systems},
volume      =   {33},
pages       =   {21464--21475},
year        =   {2020}
}

@inproceedings{
du2022vos,
title={Towards Unknown-aware Learning with Virtual Outlier Synthesis},
author={Xuefeng Du and Zhaoning Wang and Mu Cai and Sharon Li},
booktitle={International Conference on Learning Representations},
year={2022},
url={https://openreview.net/forum?id=TW7d65uYu5M}
}

@InProceedings{kirchheim2022pytorch,
author      =   {Kirchheim, Konstantin and Filax, Marco and Ortmeier, Frank},
title       =   {PyTorch-OOD: A Library for Out-of-Distribution Detection Based on PyTorch},
booktitle   =   {Proceedings of the IEEE/CVF Conference on Computer Vision and Pattern Recognition (CVPR) Workshops},
month       =   {June},
year        =   {2022},
pages       =   {4351-4360}
}

@inproceedings{ruff2018deep,
title       =   {Deep one-class classification},
author      =   {Ruff, Lukas and Vandermeulen, Robert and Goernitz, Nico and Deecke, Lucas and Siddiqui, Shoaib Ahmed and Binder, Alexander and M{\"u}ller, Emmanuel and Kloft, Marius},
booktitle   =   {International conference on machine learning},
pages       =   {4393--4402},
year        =   {2018},
organization=   {PMLR}
}

@inproceedings{wen2016discriminative,
title       =   {A discriminative feature learning approach for deep face recognition},
author      =   {Wen, Yandong and Zhang, Kaipeng and Li, Zhifeng and Qiao, Yu},
booktitle   =   {Computer vision--ECCV 2016: 14th European conference, amsterdam, the netherlands, October 11--14, 2016, proceedings, part VII 14},
pages       =   {499--515},
year        =   {2016},
organization=   {Springer}
}

@software{pytorch2019lightning,
author      =   {Falcon, William and {The PyTorch Lightning team}},
doi         =   {10.5281/zenodo.3828935},
license     =   {Apache-2.0},
month = mar,
title       =   {{PyTorch Lightning}},
URL         =   {https://github.com/Lightning-AI/lightning},
version     =   {1.4},
year        =   {2019}
}

@article{fawcett2006introduction,
title       =   {An introduction to ROC analysis},
author      =   {Fawcett, Tom},
journal     =   {Pattern recognition letters},
volume      =   {27},
number      =   {8},
pages       =   {861--874},
year        =   {2006},
publisher   =   {Elsevier}
}

@article{powers2020evaluation,
title       =   {Evaluation: from precision, recall and F-measure to ROC, informedness, markedness and correlation},
author      =   {Powers, David MW},
journal     =   {arXiv preprint arXiv:2010.16061},
year        =   {2020}
}

@inproceedings{humblot2023beyond,
title       =   {Beyond AUROC \& co. for evaluating out-of-distribution detection performance},
author      =   {Humblot-Renaux, Galadrielle and Escalera, Sergio and Moeslund, Thomas B},
booktitle   =   {Proceedings of the IEEE/CVF Conference on Computer Vision and Pattern Recognition},
pages       =   {3881--3890},
year        =   {2023}
}

@inproceedings{caron2021emerging,
title       =   {Emerging Properties in Self-Supervised Vision Transformers},
author      =   {Caron, Mathilde and Touvron, Hugo and Misra, Ishan and J\'egou, Herv\'e  and Mairal, Julien and Bojanowski, Piotr and Joulin, Armand},
booktitle   =   {Proceedings of the International Conference on Computer Vision (ICCV)},
year        =   {2021}
}

@inproceedings{chen2020simple,
title       =   {A simple framework for contrastive learning of visual representations},
author      =   {Chen, Ting and Kornblith, Simon and Norouzi, Mohammad and Hinton, Geoffrey},
booktitle   =   {International conference on machine learning},
pages       =   {1597--1607},
year        =   {2020},
organization={PmLR}
}

@article{sun2021react,
title={React: Out-of-distribution detection with rectified activations},
author={Sun, Yiyou and Guo, Chuan and Li, Yixuan},
journal={Advances in neural information processing systems},
volume={34},
pages={144--157},
year={2021}
}

@inproceedings{sun2022dice,
title={Dice: Leveraging sparsification for out-of-distribution detection},
author={Sun, Yiyou and Li, Yixuan},
booktitle={European conference on computer vision},
pages={691--708},
year={2022},
organization={Springer}
}

@inproceedings{zhang2022out,
title={Out-of-distribution detection based on in-distribution data patterns memorization with modern hopfield energy},
author={Zhang, Jinsong and Fu, Qiang and Chen, Xu and Du, Lun and Li, Zelin and Wang, Gang and Han, Shi and Zhang, Dongmei and others},
booktitle={The Eleventh International Conference on Learning Representations},
year={2022}
}

@article{shannon1948mathematical,
title={A mathematical theory of communication},
author={Shannon, Claude E},
journal={The Bell system technical journal},
volume={27},
number={3},
pages={379--423},
year={1948},
publisher={Nokia Bell Labs}
}

@book{caro2010conservation,
title={Conservation by proxy: indicator, umbrella, keystone, flagship, and other surrogate species},
author={Caro, Tim},
year={2010},
publisher={Island Press}
}

@inproceedings{williams2000flagship,
title={Flagship species, ecological complementarity and conserving the diversity of mammals and birds in sub-Saharan Africa},
author={Williams, Paul H and Burgess, Neil D and Rahbek, Carsten},
booktitle={Animal Conservation Forum},
volume={3},
number={3},
pages={249--260},
year={2000},
organization={Cambridge University Press}
}

@article{di2021pan,
title={A pan-African spatial assessment of human conflicts with lions and elephants},
author={Di Minin, Enrico and Slotow, Rob and Fink, Christoph and Bauer, Hans and Packer, Craig},
journal={Nature communications},
volume={12},
number={1},
pages={2978},
year={2021},
publisher={Nature Publishing Group UK London}
}

@article{braczkowski2023unequal,
title={The unequal burden of human-wildlife conflict},
author={Braczkowski, Alexander R and O’Bryan, Christopher J and Lessmann, Christian and Rondinini, Carlo and Crysell, Anna P and Gilbert, Sophie and Stringer, Martin and Gibson, Luke and Biggs, Duan},
journal={Communications Biology},
volume={6},
number={1},
pages={182},
year={2023},
publisher={Nature Publishing Group UK London}
}

@article{dunham2010human,
title={Human--wildlife conflict in Mozambique: a national perspective, with emphasis on wildlife attacks on humans},
author={Dunham, Kevin M and Ghiurghi, Andrea and Cumbi, Rezia and Urbano, Ferdinando},
journal={Oryx},
volume={44},
number={2},
pages={185--193},
year={2010},
publisher={Cambridge University Press}
}

@article{sedhain2016human,
  title={Human-Rhino conflict: Local people’s adaptation to impacts of rhino},
  author={Sedhain, Jyoti and Adhikary, Anukram},
  journal={Journal of Forest and Livelihood},
  volume={14},
  number={1},
  pages={53--66},
  year={2016}
}

@inproceedings{janani2022human,
title={Human-Animal Conflict Analysis and Management-A Critical Survey},
author={Janani, V and Shanthi, C},
booktitle={2022 11th International Conference on System Modeling \& Advancement in Research Trends (SMART)},
pages={1003--1007},
year={2022},
organization={IEEE}
}

@article{sambhaji2019early,
title={Early Warning System for Detection of Harmful Animals using IOT},
author={Sambhaji, Sahane Pradnya and Sanjiv, Salunke Nikita and Somnath, Shirsath Vitthal and Sanjay, Shukla Shreyas and Panhalkar, AR},
journal={International Journal of Advance Research and Innovative Ideas in Education},
volume={5},
number={3},
pages={2395--4396},
year={2019}
}

@article{bavane2018protection,
title={Protection of crops from wild animals using Intelligent Surveillance System},
author={Bavane, Vikas and Raut, Arti and Sonune, Swapnil and Bawane, AP and Jawandhiya, PM},
journal={International Journal of Research in Advent Technology (IJRAT)},
pages={2321--9637},
year={2018}
}

@inproceedings{mishra2024smart,
title={Smart Animal Repelling Device: Utilizing IoT and AI for Effective Anti-Adaptive Harmful Animal Deterrence},
author={Mishra, Akanksha and Yadav, Kamlesh Kumar},
booktitle={BIO Web of Conferences},
volume={82},
pages={05014},
year={2024},
organization={EDP Sciences}
}

@inproceedings{vojivr2023calibrated,
  title={Calibrated out-of-distribution detection with a generic representation},
  author={Voj{\'\i}{\v{r}}, Tom{\'a}{\v{s}} and {\v{S}}ochman, Jan and Aljundi, Rahaf and Matas, Ji{\v{r}}{\'\i}},
  booktitle={2023 IEEE/CVF International Conference on Computer Vision Workshops (ICCVW)},
  pages={4509--4518},
  year={2023},
  organization={IEEE}
}

@article{winkens2020contrastive,
  title={Contrastive training for improved out-of-distribution detection},
  author={Winkens, Jim and Bunel, Rudy and Roy, Abhijit Guha and Stanforth, Robert and Natarajan, Vivek and Ledsam, Joseph R and MacWilliams, Patricia and Kohli, Pushmeet and Karthikesalingam, Alan and Kohl, Simon and others},
  journal={arXiv preprint arXiv:2007.05566},
  year={2020}
}

@article{sehwag2021ssd,
  title={Ssd: A unified framework for self-supervised outlier detection},
  author={Sehwag, Vikash and Chiang, Mung and Mittal, Prateek},
  journal={arXiv preprint arXiv:2103.12051},
  year={2021}
}

@article{youden1950index,
  title={Index for rating diagnostic tests},
  author={Youden, William J},
  journal={Cancer},
  volume={3},
  number={1},
  pages={32--35},
  year={1950},
  publisher={Wiley Online Library}
}

@inproceedings{oh2020early,
  title={Early wildfire detection using convolutional neural network},
  author={Oh, Seon Ho and Ghyme, Sang Won and Jung, Soon Ki and Kim, Geon-Woo},
  booktitle={International workshop on frontiers of computer vision},
  pages={18--30},
  year={2020},
  organization={Springer}
}

@article{estiri2019semi,
  title={Semi-supervised encoding for outlier detection in clinical observation data},
  author={Estiri, Hossein and Murphy, Shawn N},
  journal={Computer methods and programs in biomedicine},
  volume={181},
  pages={104830},
  year={2019},
  publisher={Elsevier}
}

@article{scherreik2016open,
  title={Open set recognition for automatic target classification with rejection},
  author={Scherreik, Matthew D and Rigling, Brian D},
  journal={IEEE Transactions on Aerospace and Electronic Systems},
  volume={52},
  number={2},
  pages={632--642},
  year={2016},
  publisher={IEEE}
}

@inproceedings{10.1145/3674029.3674050,
author = {Seljebotn, Karoline and Lawal, Isah A.},
title = {Machine Learning Tool for Wildlife Image Classification},
year = {2024},
isbn = {9798400716379},
publisher = {Association for Computing Machinery},
address = {New York, NY, USA},
url = {https://doi.org/10.1145/3674029.3674050},
doi = {10.1145/3674029.3674050},
booktitle = {Proceedings of the 2024 9th International Conference on Machine Learning Technologies},
pages = {127–132},
numpages = {6},
location = {Oslo, Norway},
series = {ICMLT '24}
}

@misc{lila2024science,
  title = {Labeled Information Library of Alexandria: Biology and Conservationn},
  howpublished = {\url{https://lila.science/}},
}
}

\clearpage
\setcounter{page}{1}
\maketitlesupplementary
\section*{Appendix A}

\subsection{Per-Class Accuracy}

In this section we include supplementary results for the paper. The table \ref{table:BigFive} shows the accuracy performance per class of each method. To determine weather an image is In-Distribution (ID) or Out-of-Distribution (OOD) we obtain the optimal threshold using Youden's J statistic~\cite{youden1950index} similar to~\cite{oh2020early, estiri2019semi, scherreik2016open}. Youden's J quantifies the discriminative power of a binary classifier by measuring the maximum vertical distance between the Receiver Operating Characteristic (ROC) curve and the diagonal line representing a random classifier. Formally, Youden's J is formulated as:
\begin{equation}
    \mathrm{J} = \max_{\tau} (\mathrm{Sensitivity}(\tau) + \mathrm{Specificity}(\tau) - 1)
\end{equation}
where $\tau$ is the threshold, Sensitivity is the true positive rate, and Specificity is the true negative rate. The optimal threshold $\tau^{*}$ corresponds to the value of $\tau$ that maximizes J.

\subsection{Agreement Score}

We observe improved performance by introducing an agreement score between our Nearest Class Mean (NCM) and contrastive learning methods. We define two vectors $v_1, v_2 \in \mathbb{R}^n$ for $n$ classes. $v_1$ captures inference-based predictions and $v_2$ encodes feature-based information. In particular, $v_2$ is a vector of softmax values obtained from the classification head. For our NCM strategy, $v_2$ is the distance of the features of image $x$ to each class mean $\mu_c$. For contrastive learning, $v_2$ is the number of nearest neighbours for each class given the features of $x$. We normalize $v_2$ by $k$ to convert it into a probability distribution across the five target classes. Then the agreement score is given by:
\begin{equation}
    \frac{1 - \mathrm{H}(v_{1} \odot v_{2} + \epsilon)}{log(n)} \times (1-\mathrm{\mathrm{JS}(v_{1}, v_{2})})
\end{equation}
where H is the entropy and JS is the Jansen Shannon equation given by:
\begin{equation}
    \mathrm{H}(p) = - \sum_{i=1}^{n} p_{i} log p_{i}
\end{equation}
and
\begin{equation}
    \mathrm{JS}(v_{1}, v_{2}) = \frac{1}{2} [\mathrm{D}_{\mathrm{KL}}(v_{1} || m) + \mathrm{D}_{\mathrm{KL}}(v_{2} || m)]
\end{equation}
such that $\odot$ is an element-wise multiplication operation, $\epsilon \ll 1$ is a small very to prevent division by zero, $\mathrm{D}_{\mathrm{KL}}(\cdot || \cdot)$ is the Kullback-Leibler (KL) divergence and $m = \frac{1}{2} (v_1 + v_{2})$ is the midpoint distribution.
\subsection{Limitations}

The weights for SpeciesNet are in Tensorflow format. We had to use ONNX to convert them to PyTorch. We cannot guarantee the stability of the conversion process.

\begin{table*}[htbp!]
\caption{Per-class accuracy performance of each method. The top three best-performing methods are highlighted.}
\centering
\resizebox{\textwidth}{!}{%
\begin{tabular}{lrrrr|rrrr|rrrr|rrrr|rrrr|rrrr}
\toprule
 \textit{\textbf{Species}} & \multicolumn{4}{c}{\textbf{Elephant}} & \multicolumn{4}{c}{\textbf{Buffalo}} & \multicolumn{4}{c}{\textbf{Lion}} & \multicolumn{4}{c}{\textbf{Leopard}} & \multicolumn{4}{c}{\textbf{Rhino}} & \multicolumn{4}{c}{\textbf{Out-Of-Dist'}}\\
\cmidrule(lr){2-5} \cmidrule(lr){6-9} \cmidrule(lr){10-13} \cmidrule(lr){14-17} \cmidrule(lr){18-21} \cmidrule(lr){22-25} \\
\textit{\textbf{Models}} 
& \rot{\makecell[l]{Species-\\Net}} & \rot{\makecell[l]{Mega-\\Classifier}} & \rot{BioClip} & \rot{ImageNet}
& \rot{\makecell[l]{Species-\\Net}} & \rot{\makecell[l]{Mega-\\Classifier}} & \rot{BioClip} & \rot{ImageNet}
& \rot{\makecell[l]{Species-\\Net}} & \rot{\makecell[l]{Mega-\\Classifier}} & \rot{BioClip} & \rot{ImageNet}
& \rot{\makecell[l]{Species-\\Net}} & \rot{\makecell[l]{Mega-\\Classifier}} & \rot{BioClip} & \rot{ImageNet}
& \rot{\makecell[l]{Species-\\Net}} & \rot{\makecell[l]{Mega-\\Classifier}} & \rot{BioClip} & \rot{ImageNet}
& \rot{\makecell[l]{Species-\\Net}} & \rot{\makecell[l]{Mega-\\Classifier}} & \rot{BioClip} & \rot{ImageNet}\\
\bottomrule\toprule
\textbf{\textit{OOD Methods}} \\
Baseline & 
.475 & \cellcolor{blue!20}\textbf{.884} & \cellcolor{blue!20}\textbf{.827} & \cellcolor{blue!20}\textbf{.904} & \cellcolor{blue!20}{.799} & \cellcolor{blue!20}\textbf{.882} & \cellcolor{blue!20}\textbf{.889} & \cellcolor{blue!20}\textbf{.951} & \ccb{.084} & \cellcolor{blue!20}\textbf{.843} & \cellcolor{blue!20}\textbf{.761} & \cellcolor{blue!20}\textbf{.905} & \cellcolor{blue!20}.322 & \cellcolor{blue!20}\textbf{.902} & \cellcolor{blue!20}{.939} & \cellcolor{blue!20}\textbf{.958} & \cellcolor{blue!20}.114 & \cellcolor{blue!20}{.294} & \cellcolor{blue!20}{.335} & \cellcolor{blue!20}.529 & - & - & - & - \\
\textit{inference-based} \\
MaxSoftmax          & .237 & .273 & .451 & .645 & .383 & .247 & .620 & .793 & .026 & .358 & .509 & .713 & .248 & .716 & .920 & .946 & .047 & .044 & .142 & .156 & .613 & \ccb{.753} & \ccb{.721} & .718 \\
MaxLogit            & .000 & .177 & .532 & .683 & .001 & .139 & .676 & .813 & .000 & .327 & .511 & .727 & .121 & .706 & .924 & \ccb{.950} & .000 & .027 & .165 & .227 & .396 & .716 & .673 & .699 \\
Temp'Scaling  & .131 & .037 & \ccb{.801} & .677 & .327 & .077 & \ccb{.878} & .663 & .023 & .092 & \ccb{.724} & .562 & .221 & .321 & \ccb{.938} & .849 & .058 & .038 & \ccb{.329} & .351 & .613 & \ccb{.908} & .040 & .493 \\
OpenMax             & .198 & .388 & .462 & .620 & \ccb{.667} & .307 & \ccb{.712} & .799 & .078 & .499 & .440 & .672 & .293 & .782 & .916 & .945 & .044 & .107 & .172 & .240 & .506 & .554 & .658 & .696 \\
KNN                 & .036 & .098 & .390 & .563 & .137 & .090 & .526 & .525 & .010 & .079 & .436 & .518 & .001 & .014 & .837 & .892 & .041 & .037 & .236 & .360 & .641 & .637 & \ccb{.748} & .769 \\
EnergyBased         & .000 & \ccb{.689} & .497 & .752 & .000 & \ccb{.520} & .644 & \ccb{.862} & .000 & \ccb{.605} & .483 & .785 & .079 & \ccb{.837} & .921 & \ccb{.951} & .000 & \ccB{.331} & .164 & .300 & .403 & .668 & .701 & .693 \\
Entropy             & .208 & .371 & .461 & .650 & .279 & .314 & .635 & .797 & .023 & .406 & .511 & .714 & .227 & .720 & .921 & .946 & .034 & .064 & .143 & .156 & \ccB{.718} & .658 & .713 & .716 \\
SHE                 & .156 & .001 & .300 & .632 & .496 & .005 & .545 & .782 & .048 & .002 & .346 & .618 & .268 & .410 & .906 & .927 & .028 & .003 & .192 & .393 & .538 & \ccB{.986} & .722 & .726 \\
ReAct               & .000 & \ccb{.459} & .524 & \ccb{.818} & .000 & .293 & .656 & \ccb{.894} & .000 & .472 & .301 & .689 & .077 & .786 & .910 & .944 & .000 & .103 & .264 & \ccB{.666} & \ccb{.694} & .688 & \ccb{.643} & .677 \\
DICE                & .000 & .208 & .405 & .685 & .000 & .091 & .568 & .818 & .000 & .344 & .423 & .755 & .082 & .719 & \ccB{.950} & .754 & .000 & .030 & .155 & .251 & .283 & .692 & .700 & .713 \\
\textit{feature-based} \\
Deep SVDD  & \ccB{.558} & .442 & .558 & .747 & \ccB{.829} & .489 & .538 & .609 & \ccb{.368} & .442 & .374 & .570 & \ccb{\textbf{.576}} & .291 & .625 & .837 & \ccb{.259} & .152 & .254 & .339 & .001 & .522 & .491 & .471 \\
Center Loss         & .511 & .275 & .537 & .630 & .837 & .389 & .639 & .811 & .187 & .454 & .357 & .602 & \ccb{.494} & .810 & .894 & .915 & .244 & .094 & .172 & .196 & .000 & .603 & .698 & .732 \\
\textit{inference and feature based} \\
GROOD & \ccb{.476} & .378 & .420 & .630 & .455 & .403 & .509 & .693 & \ccB{.420} & .414 & .483 & .663 & .409 & .503 & .553 & .784 & \ccB{.406} & \ccB{.393} & \ccB{.402} & \ccb{.530} & .554 & .683 & \ccB{.853} & \ccB{.893} \\
\midrule
\textit{proposed baselines} \\
\textbf{NCM Agreement}                  & .171 & .174 & .610 & .699 & .378 & .364 & .686 & .831 & .015 & \ccb{.606} & \ccb{.655} & \ccb{.820} & .296 & .817 & .891 & .928 & .071 & .196 & .194 & .523 & \ccb{.654} & .621 & .509 & \ccb{.887} \\
\textbf{Contrastive Loss}    & \ccb{.488} & .456 & \ccb{.686} & \ccb{.844} & .651 & .409 & .688 & .856 & .078 & .488 & .591 & \ccb{.808} & .490 & \ccb{.849} & .910 & 944 & .137 & .034 & .232 & .419 & .562 & \ccb{.544} & .599 & .723 \\
\textbf{NCM Agreement Score} & .337 & .333 & .460 & .607 & .546 & .271 & .642 & .642 & .053 & .361 & .544 & .522 & .279 & .691 & .871 & .870 & .060 & .132 & .060 & .286 & .444 & .564 & .714 & \ccb{.872} \\
\textbf{Contrastive Agreement Score} & 346 & .390 & .438 & .634 & .526 & \ccb{.584} & .520 & .699 & .011 & .043 & .462 & .748 & .224 & .456 & .901 & .916 & .007 & .092 & .143 & .303 & .525 & .589 & .691 & .649 \\
\bottomrule
\end{tabular}
}
\label{table:BigFive}
\end{table*}

\end{document}